\documentclass[11pt]{arxform}
\usepackage{graphicx}
\usepackage{multirow}
\usepackage{todonotes}
\usepackage{stmaryrd}
\usepackage{amsmath}
\usepackage{amssymb}
\usepackage{hyperref}
\usepackage{subfig}
\usepackage{algorithm}
\usepackage{algorithmic}
\usepackage{xcolor}
\usepackage{hyperref}

\usepackage{times}

\newcommand{\given}{\, | \,}

\renewcommand{\vec}[1]{\boldsymbol{#1}}
\newcommand*{\defeq}{\mathrel{\vcenter{\baselineskip0.5ex \lineskiplimit0pt
			\hbox{\footnotesize.}\hbox{\footnotesize.}}}%
	=}

\newcommand{\fromto}{\longrightarrow}

\newcommand{\cX}{\mathcal{X}}
\newcommand{\cL}{\mathcal{L}}
\newcommand{\cY}{\mathcal{Y}}
\newcommand{\cR}{\mathcal{R}}
\newcommand{\cD}{\mathcal{D}}
\newcommand{\bx}{\boldsymbol{x}}
\newcommand{\by}{\boldsymbol{y}}

\newcommand{\bh}{\boldsymbol{h}}

\newcommand{\br}{\boldsymbol{r}}
\newcommand{\loss}{\ell}

\begin{document}

\title{Conformal Rule-Based Multi-label Classification\thanks{Draft of an article presented at KI 2020, 43.\ German Conference on Artificial Intelligence, Bamberg, Germany}}
%
%

\author{Eyke H\"ullermeier\inst{1} \and Johannes F\"urnkranz\inst{2} \and Eneldo Loza Mencia\inst{3}}
\institute{Paderborn University, Paderborn, Germany
\and
Johannes Kepler University, Linz, Austria
\and
Technical University Darmstadt, Germany}

%
\maketitle             

\begin{abstract}
We advocate the use of conformal prediction (CP) to enhance rule-based multi-label classification (MLC). In particular, we highlight the mutual benefit of CP and rule learning: Rules have the ability to provide natural (non-)conformity scores, which are required by CP, while CP suggests a way to calibrate the assessment of candidate rules, thereby supporting better predictions and more elaborate decision making.
We illustrate the potential usefulness of calibrated conformity scores in a case study on lazy multi-label rule learning.
\end{abstract}

\section{Introduction}

The setting of multi-label classification (MLC), which generalizes standard multi-class classification by relaxing the assumption of mutual exclusiveness of classes, has received a lot of attention in machine learning, and various methods for tackling this problem have been proposed in the literature \cite{Zhang:2014}. A \emph{rule-based approach} to MLC is appealing and comes with a number of interesting properties. For example, rules are potentially interpretable and can provide explanations of a prediction \cite{loza2018}. Moreover, due to their local nature, rule-based predictors are very expressive and can adapt to local properties of the data in a flexible way. 

In the context of MLC, the local nature of rules may also cause difficulties, however. In particular, due to the imbalance between positive and negative labels, which is typical for MLC, ``good'' rules with positive predictions that can stand up to negative rules are difficult to find. Here, we advocate the combination of multi-label rule learning with conformal prediction (CP) to mitigate this problem. To the best of our knowledge, CP has not been used in the context of MLC (neither rule-based nor otherwise) so far.

\section{Multilabel Classification}

Let $\cX$ denote an instance space, and let $\cL= \{\lambda_k \}_{k=1}^K$ be a finite set of class labels. We assume that an instance $\bx \in \cX$ is (probabilistically) associated with a subset of labels $\Lambda = \Lambda(\bx) \in 2^\cL$; this subset is often called the set of relevant (positive) labels, while the complement $\cL \setminus \Lambda$ is considered as irrelevant (negative) for $\bx$. We identify a set $\Lambda$ of relevant labels with a binary vector $\by = (y_1, \ldots, y_K)$, where $y_k = \llbracket \lambda_k \in \Lambda \rrbracket$.\footnote{$\llbracket \cdot \rrbracket$ is the indicator function, i.e., $\llbracket A \rrbracket = 1$ if the predicate $A$ is true and $=0$ otherwise.} By $\cY = \{0,1\}^K$ we denote the set of possible labelings.

Given training data 
$\cD = \{ (\bx_n,\by_n) \}_{n=1}^N  \subset \cX \times \cY$,
the goal in MLC is to learn a predictive model in the form of a multilabel classifier $\bh$, which is a mapping $\cX \fromto \cY$ that assigns a (predicted) label subset to each instance $\bx\in \cX$. Thus, the output of a classifier $\bh$ is a vector of predictions $\bh(\bx) = (h_1(\bx), \ldots , h_K(\bx)) \in \{ 0,1 \}^K$, also denoted as $\hat{\by} = (\hat{y}_1, \ldots , \hat{y}_K)$. For measuring the (generalization) performance of such a model, a large spectrum of loss functions or performance metrics have been proposed in the literature, including the Hamming loss
$\loss_H(\by, \hat{\by}) \defeq \frac{1}{K}
\sum_{k=1}^K  \, \llbracket y_k \neq  \hat{y}_k \rrbracket$
and the F1-measure \cite{mpub230}.

\section{Conformal Prediction}

Conformal prediction \cite{vovk_al,shaf_at08,bala_cp,gamm_cp19} is a framework for reliable prediction that is rooted in classical frequentist statistics and hypothesis testing. Given a sequence of training observations 
$$
(\vec{x}_1, y_1), \,  (\vec{x}_2, y_2), \ldots ,  (\vec{x}_N, y_N), \, (\vec{x}_{N+1}, \bullet)
$$ 
and a new query $\vec{x}_{N+1}$ with unknown outcome $y_{N+1}$,
the basic idea is to hypothetically replace $\bullet$ by each candidate, i.e., to test the hypothesis $y_{N+1} = y$ for all $y \in \mathcal{Y}$. Only those outcomes $y$ for which this hypothesis can be rejected at a predefined level of confidence are excluded, while those for which the hypothesis cannot be rejected are collected to form the prediction set or \emph{prediction region} $Y \subseteq \mathcal{Y}$. By construction, the set-valued prediction $Y = Y(\vec{x}_{n+1})$ is guaranteed to cover the true outcome $y_{N+1}$ with a pre-specified probability of $1- \epsilon$ (for example 95\,\%).

Hypothesis testing is done in a nonparametric way: Consider any ``nonconformity'' function $f: \, \mathcal{X} \times \mathcal{Y} \longrightarrow \mathbb{R}$ that assigns scores $\alpha = f(\vec{x}, y)$ to input/output tuples; the latter can be interpreted as a measure of ``strangeness'' of the pattern $(\vec{x}, y)$, i.e., the higher the score, the less the data point $(\vec{x}, y)$ conforms to what one would expect to observe. Applying this function to the sequence of observations, with a specific (though hypothetical) choice of $y = y_{N+1}$, yields a sequence of scores
$\alpha_1, \, \alpha_2, \ldots , \alpha_N , \, \alpha_{N+1}$, where $\alpha_i = f(\vec{x}_i, y_i)$.
Denote by $\sigma$ the permutation of $\{1, \ldots , N+1\}$ that sorts the scores in increasing order, i.e., such that $\alpha_{\sigma(1)} \leq \ldots \leq \alpha_{\sigma(N+1)}$. Under the assumption that the hypothetical choice of $y_{N+1}$ is in agreement with the true data-generating process, and that this process has the property of exchangeability (which is weaker than the assumption of independence and essentially means that the order of observations is irrelevant), every permutation $\sigma$ has the same probability of occurrence. Consequently, the probability that $\alpha_{N+1}$ is among the $\epsilon$\,\% highest nonconformity scores should be low. This notion can be captured by the $p$-values associated with the candidate $y$, defined as 
\begin{equation}\label{eq:pvalue}
p(y) \defeq \frac{\# \{ i \given \alpha_i \geq \alpha_{N+1} \}}{N+1}
\end{equation}
According to what we said, the probability that $p(y) < \epsilon$ (i.e., $\alpha_{N+1}$ is among the $\epsilon$\,\% highest $\alpha$-values) is upper-bounded by $\epsilon$. 
Thus, the hypothesis $y_{N+1} = y$ can be rejected for those candidates $y$ for which $p(y) < \epsilon$. 

Conformal prediction as outlined above realizes transductive inference, although inductive variants also exist \cite{papa_ic08}, where the nonconformity scores in (\ref{eq:pvalue}) are produced on a training resp.\ validation data set. The error bounds are valid and well calibrated by construction, regardless of the nonconformity function $f$. However, the choice of this function has an important influence on the \emph{efficiency} of conformal prediction, that is, the size of prediction regions: The more suitably the nonconformity function is chosen, the smaller these sets will be.

\section{Conformal Rule-Based MLC}

A rule-based classifier in the context of MLC is understood as a collection $\cR = \{ \br_1, \ldots , \br_M \}$ of individual rules $\br_m$, where each rule $\br_m:\, H_m \leftarrow B_m$ is characterized by a \emph{head} $H_m$ and a \emph{body} $B_m$. Roughly speaking, the rule head makes an assertion about the relevance of the labels $\lambda_k$, 
while the rule body specifies conditions under which this assertion is valid. It typically appears in the form of a logical predicate that specifies conditions on a query instance $\bx$, for example a logical conjunction of restrictions on some of the features (e.g., a numerical value must lie in a certain interval).

\subsection{Lazy Rule Learning}
  
Here, we consider a \emph{lazy} approach to multi-label rule learning, in which, instead of (eagerly) inducing a complete model $\cR$ from the training data $\cD$, a single rule $\br_q:\, H_q \leftarrow B_q$ is induced at prediction time \cite{aha_ll,frie_ld96}. This rule is specifically tailored to a query instance $\bx_q$, for which a prediction is sought. More concretely, considering a binary relevance approach, a separate rule $\br_{q,k}:\, H_{q,k} \leftarrow B_{q,k}$ is constructed for each label $\lambda_k \in \cL$. The rule head is of the form $\hat{y}_k = 0$ or $\hat{y}_k = 1$. In the first case, the rule is a negative rule that predicts $\lambda_k$ to be irrelevant, in the second case a positive rule that predicts $\lambda_k$ to be relevant.  

The local nature of rules has advantages but may also cause difficulties, especially in the context of MLC, where the data is highly imbalanced. In many cases, only a tiny fraction of the labels is relevant (positive), while the majority is irrelevant (negative). In general, this makes it difficult to find a ``good'' rule with positive predictions in its head, where the quality of a rule is typically measured in terms of two criteria, namely support (the body should be general enough so as to cover many instances) and confidence (the covered instances should belong to the same class). On the contrary, the learner has a strong incentive to make negative predictions, especially for loss functions such as Hamming. For example, the default rule with empty body, which predicts all labels to be always negative, will often have a very low Hamming loss, because most labels will be negative in the test examples. At the same time, this rule has a large support. When learning a single rule, as opposed to a complete model with many rules, that single rule must at least be better than the default rule\,---\,which is difficult for positive rules, as these normally have a small support.    

\subsection{Conformity of Positive and Negative Predictions}

In general, the evaluation of negative rules is systematically better than the evaluation of positive rules. This is a motivation for the use of conformal prediction, which, if applied in a per-class manner, could ``calibrate'' the evaluations. More specifically, for a query instance $\bx_q$ and a label $\lambda_k \in \cL$, we propose the conformity (instead of non-conformity) score 
\begin{equation}\label{eq:conf}
c(\bx_q , y_k ) \defeq \max_{\br \in C(\bx_q , y_k)} eval( \br) \, ,
\end{equation}
where $y_k \in \{ 0 , 1  \}$, $C(\bx_q , y_k)$ is a set of candidate rules that cover $\bx_q$ and predict $y_k$ for the label $\lambda_k$, and $eval$ is an evaluation measure informing about the quality of the rule $\br$. As already said, such measures typically depend on the confidence and the support of the rule. In our illustration below, we shall use the lower confidence bound $\hat{p} - \sqrt{1/n}$, where $n$ is the number of examples covered by the rule and $\hat{p}$ the fraction of examples with the predicted label \cite{auer_ft02}, though any other measure could be used as well. Practically, it might be difficult to determine the maximum in (\ref{eq:conf}) exactly, as an exhaustive search of the candidate set $C(\bx_q , y_k)$ might be infeasible. Instead, greedy search techniques are often used to find an approximately optimal rule.

The measure (\ref{eq:conf}) appears to be a very natural measure of conformity: The conformity of $y_k$ for $\bx_q$ is high if a high-quality rule can be found that predicts $y_k$. A measure of \emph{plausibility} of this label is then given by
\begin{equation}\label{eq:plaus}
q(\bx_q , y_k ) = 1 - p(\bx_q , y_k ) =  \frac{ \# \big\{ (\bx , y) \in \cD \given  y= y_k , c(\bx_q , y_k ) > c(\bx , y) \big\}  }{\# \big\{ (\bx , y) \in \cD \given  y= y_k  \big\}} \, ,
\end{equation}
where $\cD$ is the training data
and $c(\bx , y)$ the conformity of the training example $(\bx , y)$ determined in a leave-one-out manner (i.e., the quality of the best rule for $(\bx , y)$ found in $\cD \setminus \{ (\bx , y) \}$). In other words, if $q(\bx_q , 1 ) = \alpha$, it means that the quality of the best positive rule for $\bx_q$ is better than the quality of $100\, \alpha \%$ of the rules found for the truly positive examples in the training data, and the same interpretation applies to $q(\bx_q , 0 )$. Consequently, only low values close to 0 provide real evidence \emph{against} a certain prediction. For example, if $q(\bx_q , 1 ) = 0.2$, it means that the positive rule found for $\bx_q$ is still better than $20\%$ of the rules for the truly positive examples in the training data. In the spirit of hypothesis testing, one would ``reject'' the positive class only if $q(\bx_q , 1 ) < t$ for some critical threshold $t$ such as $t=0.1$ or $t=0.05$, and similarly for the negative class.


\begin{figure}
    \centering
     \includegraphics[width=0.45\linewidth]{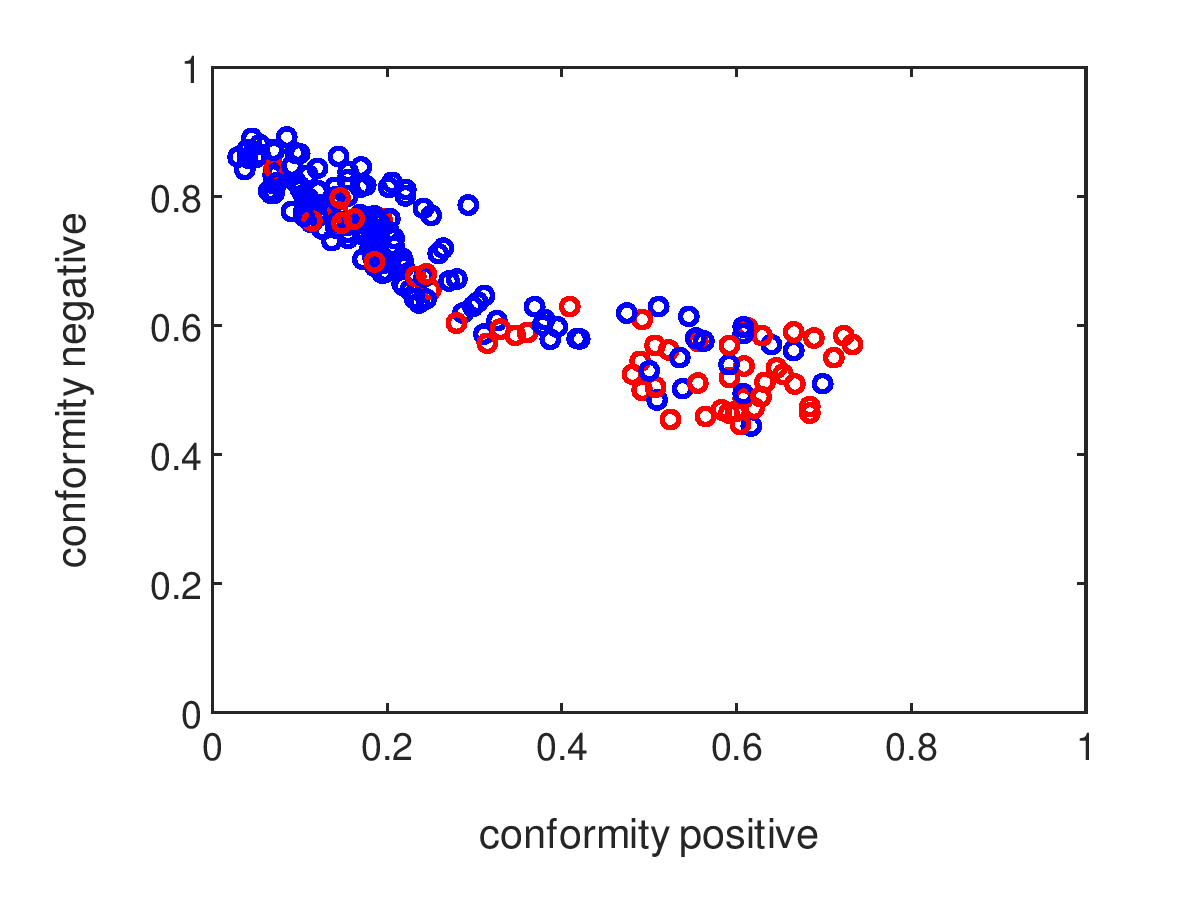}
    \includegraphics[width=0.45\linewidth]{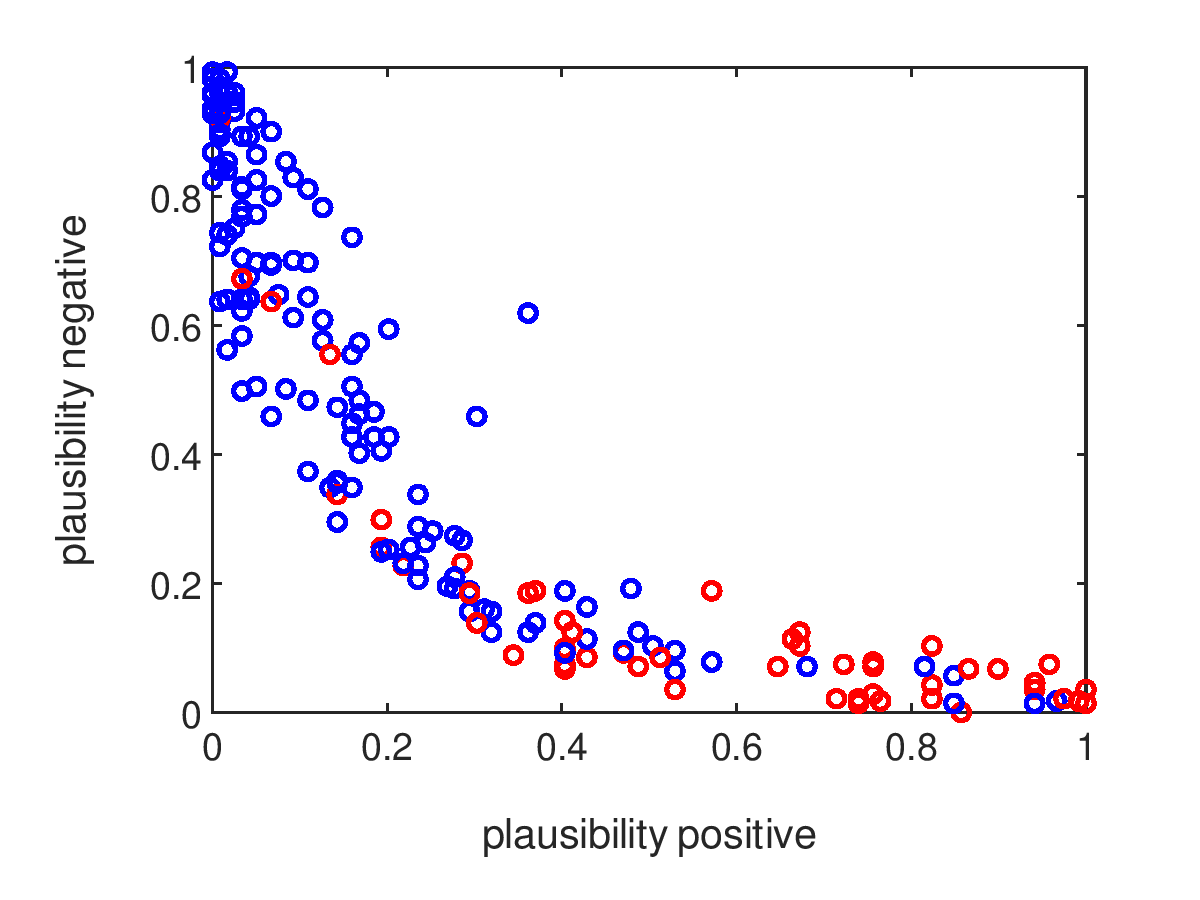}
       \vspace*{-4mm}
    \caption{Positive and negative conformity scores (\ref{eq:conf}) and calibrated plausibilities (\ref{eq:plaus}) for the first label in the emotions data. Positive examples are plotted as red, negative examples as blue points.}
    \label{fig:distr}
\end{figure}

As an illustration, Fig.\ \ref{fig:distr} shows the distribution of positive and negative 
conformity scores (\ref{eq:conf}) and calibrated plausibilities (\ref{eq:plaus}) 
for the first label in the emotions data (on a randomly chosen training set of size 400), a common benchmark data set with 596 examples, 72 attributes, and 6 labels \cite{wiec_ml06}. Here, simple rules in the form of Parzen windows \cite{parz_oe62} have been learned, searching the space of such rules in a greedy, bottom-up manner (starting with a small window around $\bx_q$ and successively increasing its size). As expected, the positive examples tend to have a higher positive than negative plausibility, and vice versa for the negative examples. Moreover, the sum of the two scores tends to be upper-bounded by 1 and sometimes takes values closer to 0, suggesting higher certainty in the true label in some cases and less in others, again confirming the appropriateness of the conformity measure (\ref{eq:conf}).

\subsection{Prediction and Decision Making}

\begin{figure}
    \centering
      \includegraphics[width=0.4\linewidth]{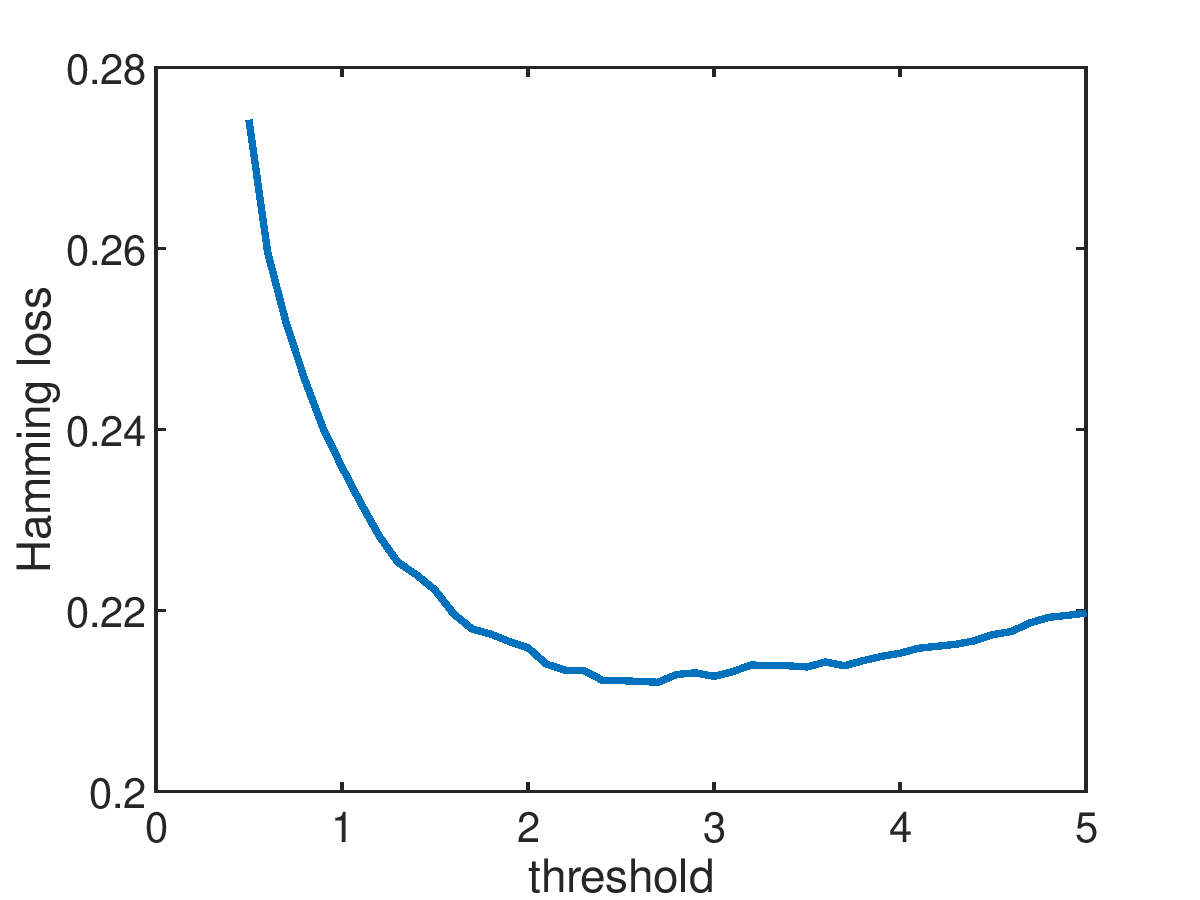}
    \includegraphics[width=0.4\linewidth]{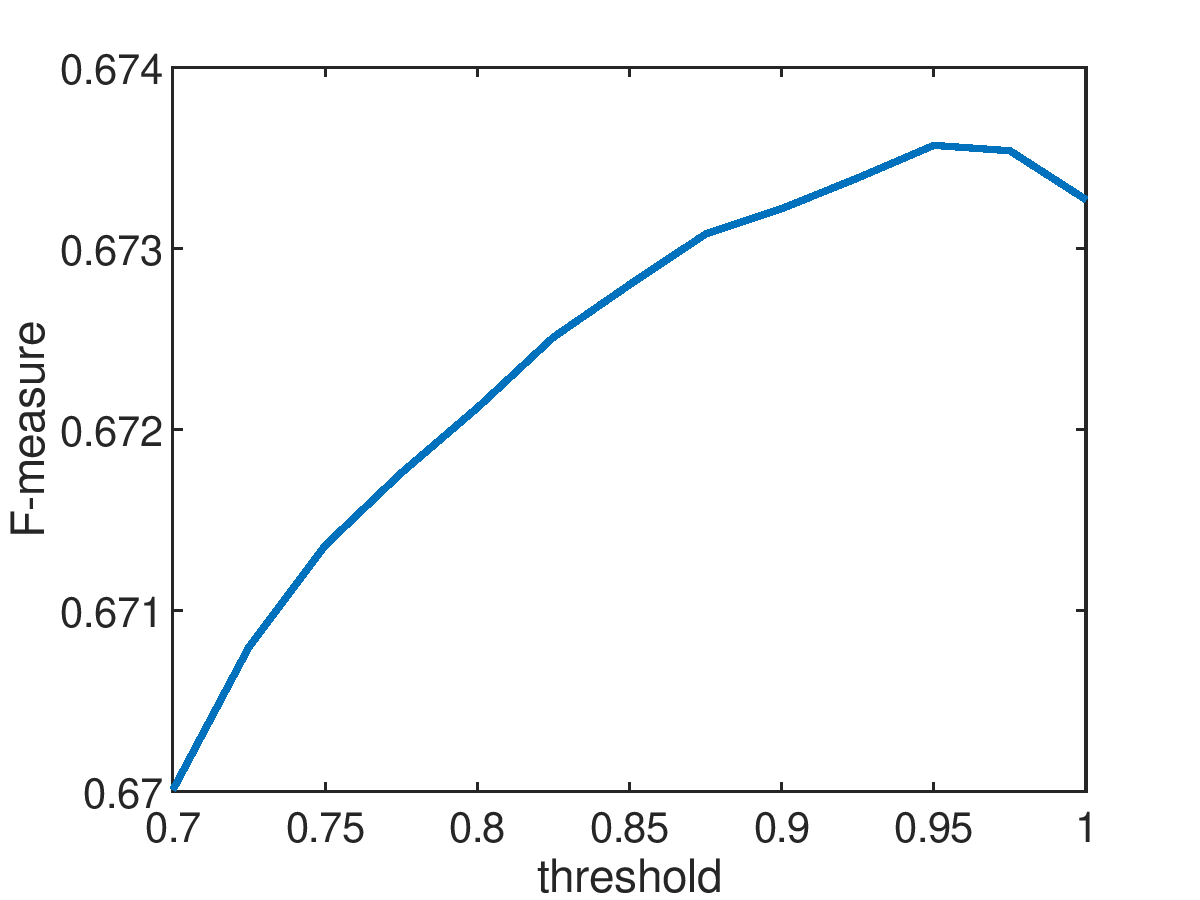}
      \includegraphics[width=0.4\linewidth]{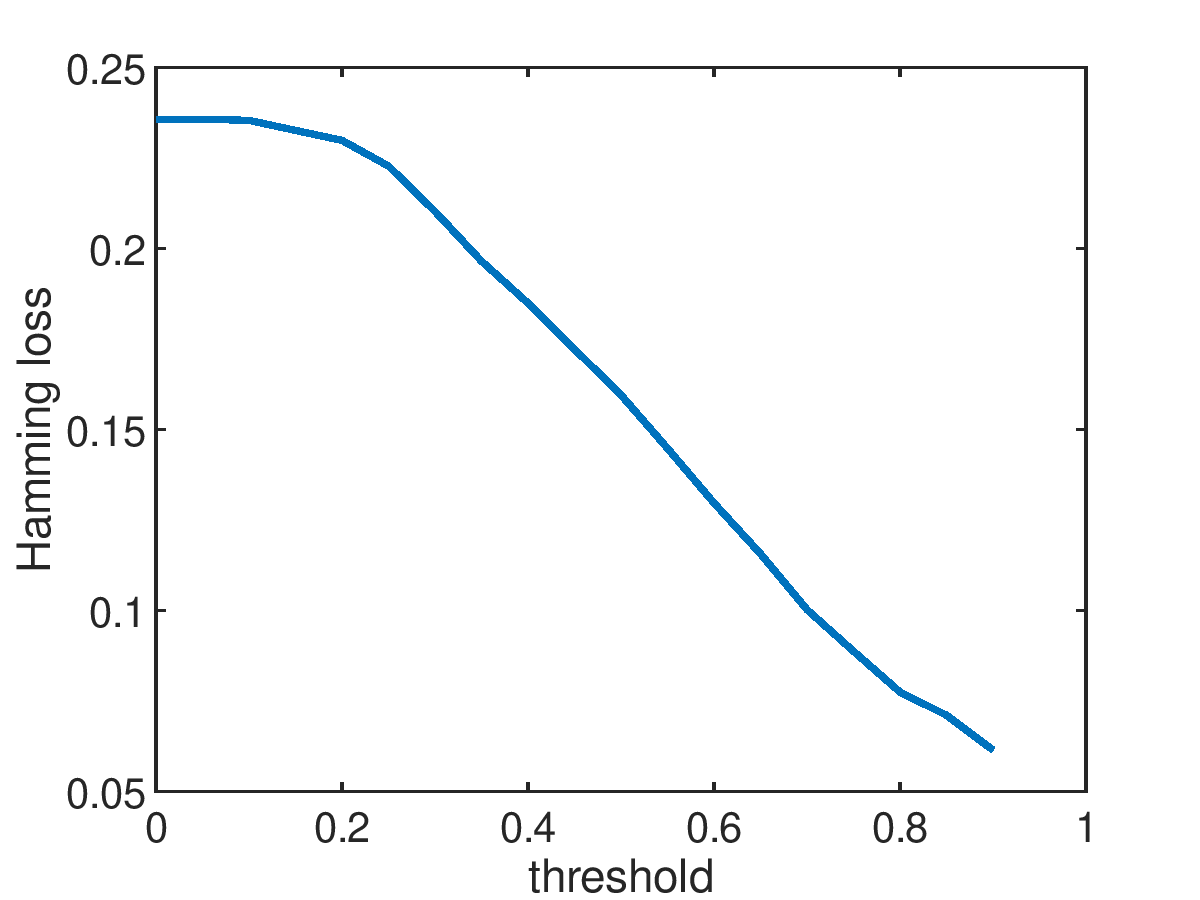}
    \includegraphics[width=0.4\linewidth]{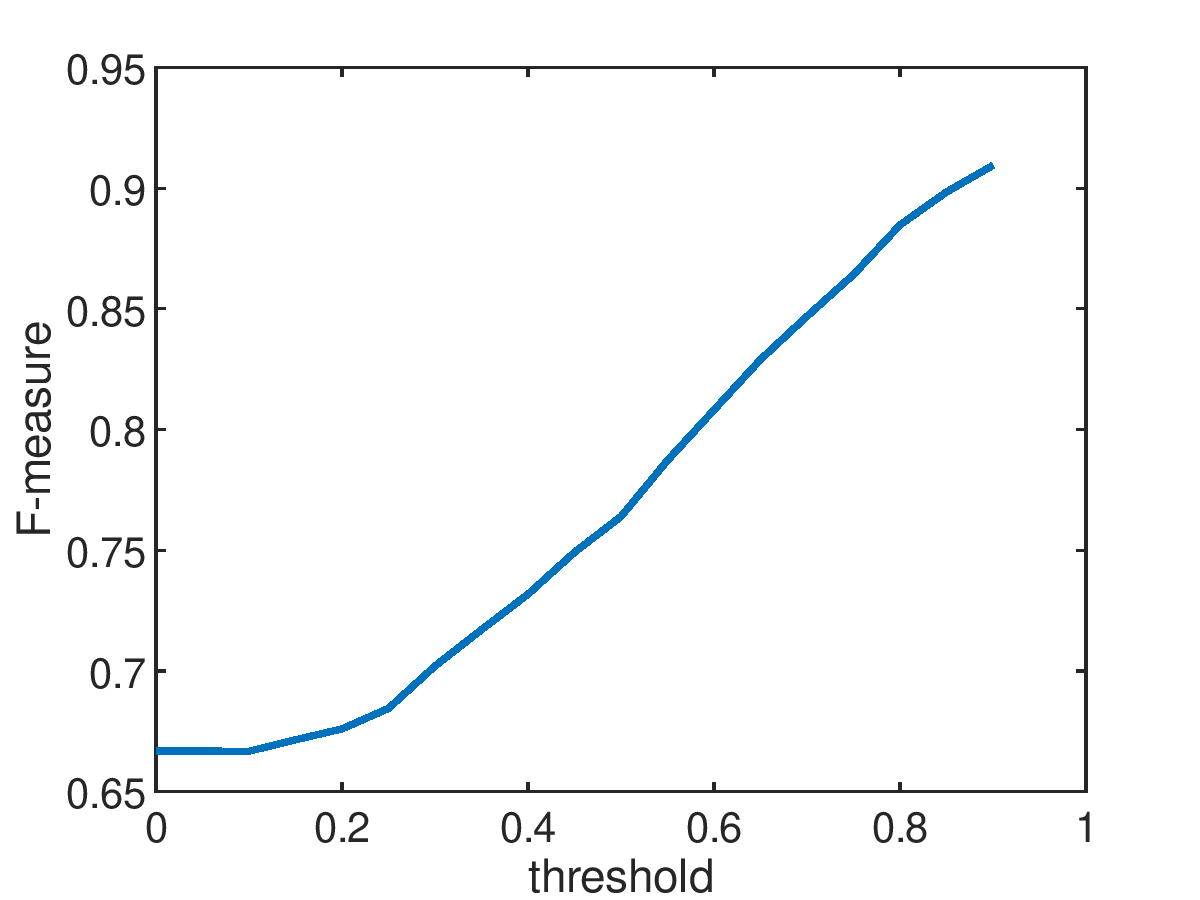}
    \vspace*{-3mm}
    \caption{Top: Hamming loss and F-measure on the emotions data, depending on the threshold $\theta$ in the decision rule (\ref{eq:threshold}). Bottom: Accuracy-rejection curves for Hamming loss and F1-measure on the same data.}
    \label{fig:threshold}
\end{figure}

Given a query $\bx_q$, the degrees $q(\bx_q , 1 )$ and $q(\bx_q , 0 )$ provide useful information about the plausibility of the positive and negative class, respectively, and hence a suitable basis for prediction and decision making. The arguably most obvious idea is to compare the two degrees and predict the label with higher plausibility, i.e., positive if $q(\bx_q , 1 ) \geq q(\bx_q , 0 )$ and negative otherwise. Yet, since MLC losses are not necessarily symmetric, and the class distribution is imbalanced, one may also think of a more general decision rule of the form 
\begin{equation}\label{eq:threshold}
\hat{y}_k = \big\llbracket q(\bx_q , 1 ) \geq \theta \cdot q(\bx_q , 0 ) \big\rrbracket \, ,
\end{equation}
where $\theta > 0$ is a parameter. Fig.\ \ref{fig:threshold} (top) shows the average test performance\footnote{50 random splits into 400 training examples and 196 test examples.}  on the emotions data in terms of the Hamming loss and (micro) F1-measure. As can be seen, by tuning the threshold $\theta$, the performance can indeed be optimized, although $\theta = 1$ is already close to optimal, confirming that the scores (\ref{eq:plaus}) are already well calibrated.

Recalling that conformal prediction is actually conceived for \emph{set-valued} prediction, one may also think of using the two plausibilities to support more sophisticated decision making. One example is multi-label classification with (partial) abstention, where the learner is allowed to abstain on those labels on which it is not certain enough \cite{nguyen20}. A natural reason to abstain, for example, is a low support for both options:
$\max\{ q(\bx_q , 0 ) , q(\bx_q , 1  ) \} \leq \theta$,
where $\theta$ is again a threshold. The effectiveness of such an approach is shown by the accuracy-rejection curves in Fig.\ \ref{fig:threshold} (bottom), which depict the average Hamming loss and F1-measure on those parts of the test data on which the learner does not abstain. The curves show a drastic increase in performance with an increasing amount of abstention (i.e., increasing $\theta$), suggesting that the learner is indeed abstaining on the right labels, namely those that are most uncertain\footnote{Note that the accuracy-rejection curve for random abstention is flat.}.

\section{Conclusion and Outlook}

The purpose of this paper is to highlight the potential usefulness of combining multi-label (rule) learning with conformal prediction. On the one side, rules provide a natural means for producing conformity scores of candidate labelings, very much like nearest neighbor methods, which are commonly used for CP \cite{papa_rc11}. On the other side, CP allows for producing meaningful and better calibrated measures of support in favor or label relevance, thus providing the basis for improved prediction, especially in advanced settings like MLC with abstention. 

Exploiting the potential of this approach requires answers to a multitude of questions. One important building block, for example, is the class of candidate rules $C(\bx_q,y)$ and the search in this class. Lazy rule learning as well as ensemble methods appear to be appealing in this regard. Moreover, to capture correlations and dependencies between different labels, the approach should be generalized toward the learning of rules with multi-label heads, predicting complete label combinations instead of individual labels.

\subsection*{Acknowledgements}
This work was supported by the German Research Foundation (DFG) under grant number 400845550.


\begin{thebibliography}{10}
\providecommand{\url}[1]{\texttt{#1}}
\providecommand{\urlprefix}{URL }
\providecommand{\doi}[1]{https://doi.org/#1}

\bibitem{aha_ll}
Aha, D. (ed.): Lazy Learning. Kluwer Academic Publ. (1997)

\bibitem{auer_ft02}
Auer, P., Cesa-Bianchi, N., Fischer, P.: Finite-time analysis of the multiarmed
  bandit problem. Machine Learning  \textbf{47}(2--3),  235--256 (2002)

\bibitem{bala_cp}
Balasubramanian, V., Ho, S., Vovk, V. (eds.): Conformal Prediction for Reliable
  Machine Learning: Theory, Adaptations and Applications. Morgan Kaufmann
  (2014)

\bibitem{mpub230}
Dembczynski, K., Waegeman, W., Cheng, W., H\"ullermeier, E.: On label
  dependence and loss minimization in multi-label classification. Machine
  Learning  \textbf{88}(1--2),  5--45 (2012)

\bibitem{frie_ld96}
Friedman, J., Kohavi, R., Yun, Y.: Lazy decision trees. In: Proceedings {\sc
  AAAI--96}. pp. 717--724. Morgan Kaufmann, Menlo Park, California (1996)

\bibitem{gamm_cp19}
Gammerman, A., Vovk, V., Bostr\"om, H., Carlsson, L.: Conformal and
  probabilistic prediction with applications: {E}ditorial. Machine Learning
  \textbf{108}(3),  379--380 (2019)

\bibitem{loza2018}
{Loza Mencia}, E., F\"urnkranz, J., H\"ullermeier, E., Rapp, M.: Learning
  interpretable rules for multi-label classification. In: Escalante, H.J.,
  Escalera, S., Guyon, I., Baro, X., G\"uclü\"ut\"urk, Y., G\"ucl\"u, U., van
  Gerven, M. (eds.) Explainable and Interpretable Models in Computer Vision and
  Machine Learning, pp. 81--113. The Springer Series on Challenges in Machine
  Learning, Springer-Verlag (2018)

\bibitem{nguyen20}
Nguyen, V.L, H{\"{u}}llermeier, E.: Reliable multi-label classification:
  {P}rediction with partial abstention. In: Proc.\ AAAI-20, Thirty-Fourth AAAI
  Conference on Artificial Intelligence. New York, USA (2020)

\bibitem{papa_ic08}
Papadopoulos, H.: Inductive conformal prediction: Theory and application to
  neural networks. Tools in Artificial Intelligence  \textbf{18}(2),  315--330
  (2008)

\bibitem{papa_rc11}
Papadopoulos, H., Vovk, V., Gammerman, A.: Regression conformal prediction with
  nearest neighbours. Journal of Artificial Intelligence Research  \textbf{40},
   815--840 (2011)

\bibitem{parz_oe62}
Parzen, E.: On estimation of a probability density function and mode. Annals of
  Mathematical Statistics  \textbf{33},  1065--1076 (1962)

\bibitem{shaf_at08}
Shafer, G., Vovk, V.: A tutorial on conformal prediction. Journal of Machine
  Learning Research pp. 371--421 (2008)

\bibitem{vovk_al}
Vovk, V., Gammerman, A., Shafer, G.: Algorithmic Learning in a Random World.
  Springer-Verlag (2003)

\bibitem{wiec_ml06}
Wieczorkowska, A., Synak, P., Ras, Z.: Multi-label classification of emotions
  in music. In: Klopotek, M., Wierzchon, S., Trojanowski, K. (eds.) Intelligent
  Information Processing and Web Mining. Springer, Berlin, Heidelberg (2006)

\bibitem{Zhang:2014}
Zhang, M.L., Zhou, Z.H.: A review on multi-label learning algorithms. IEEE
  Transactions on Knowledge and Data Engineering  \textbf{26}(8),  1819--1837
  (2014)

\end{thebibliography}

\end{document}